\documentclass[11pt,a4paper]{article}
\usepackage[hyperref]{acl2020}
\usepackage{times}
\usepackage{latexsym}
\usepackage{graphicx}
\usepackage{multirow}
\usepackage{lingmacros}
\usepackage{rotating}

\usepackage{microtype}

\aclfinalcopy 

\title{SemEval-2021 Task 1: Lexical Complexity Prediction}

\author{Matthew Shardlow\textsuperscript{1}, Richard Evans\textsuperscript{2}, Gustavo Henrique Paetzold\textsuperscript{3}, Marcos Zampieri\textsuperscript{4} \\ \textsuperscript{1}Manchester Metropolitan University, UK \\ \textsuperscript{2}University of Wolverhampton, UK \\  
\textsuperscript{3}Universidade Tecnológica Federal do Paraná, Brazil \\
\textsuperscript{4}Rochester Institute of Technology  USA \\
  \texttt{m.shardlow@mmu.ac.uk}
}

\date{}

\begin{document}

\maketitle

\begin{abstract}
This paper presents the results and main findings of SemEval-2021 Task 1 - Lexical Complexity Prediction. We provided participants with an augmented version of the CompLex Corpus \cite{shardlow-etal-2020-complex}. CompLex is an English multi-domain corpus in which words and multi-word expressions (MWEs) were annotated with respect to their complexity using a five point Likert scale. SemEval-2021 Task 1 featured two Sub-tasks: Sub-task 1 focused on single words and Sub-task 2 focused on MWEs. The competition attracted 198 teams in total, of which 54 teams submitted official runs on the test data to Sub-task 1 and 37 to Sub-task 2. 
\end{abstract}

\section{Introduction}
\label{intro}

The occurrence of an unknown word in a sentence can adversely affect its comprehension by readers. Either they give up, misinterpret, or plough on without understanding.
A committed reader may take the time to look up a word and expand their vocabulary, but even in this case they must leave the text, undermining their concentration. The natural language processing solution is to identify candidate words in a text that may be too difficult for a reader \cite{shardlow-2013-comparison,paetzold-specia:2016:SemEval1}. Each potential word is assigned a judgment by a system to determine if it was deemed `complex' or not. These scores indicate which words are likely to cause problems for a reader. The words that are identified as problematic can be the subject of numerous types of intervention, such as direct replacement in the setting of lexical simplification \cite{gooding2019recursive}, or extra information being given in the context of explanation generation \cite{rello2015plug}.

Whereas previous solutions to this task have typically considered the Complex Word Identification (CWI) task \cite{paetzold-specia:2016:SemEval1,stajner-EtAl:2018:BEA} in which a binary judgment of a word's complexity is given (i.e., is a word complex or not?), we instead focus on the Lexical Complexity Prediction (LCP) task \cite{shardlow-etal-2020-complex} in which a value is assigned from a continuous scale to identify a word's complexity (i.e., how complex is this word?). We ask multiple annotators to give a judgment on each instance in our corpus and take the average prediction as our complexity label. The former task (CWI) forces each user to make a subjective judgment about the nature of the word that models their personal vocabulary. Many factors may affect the annotator's judgment including their education level, first language, specialism or familiarity with the text at hand. The annotators may also disagree on the level of difficulty at which to label a word as complex. One annotator may label every word they feel is above average difficulty, another may label words that they feel unfamiliar with, but understand from the context, whereas another annotator may only label those words that they find totally incomprehensible, even in context. Our introduction of the LCP task seeks to address this annotator confusion by giving annotators a Likert scale to provide their judgments. Whilst annotators must still give a subjective judgment depending on their own understanding, familiarity and vocabulary --- they do so in a way that better captures the meaning behind each judgment they have given. By aggregating these judgments we have developed a dataset that contains continuous labels in the range of 0--1 for each instance. This means that rather than a system predicting whether a word is complex or not (0 or 1), instead a system must now predict where, on our continuous scale, a word falls (0--1).

Consider the following sentence taken from a biomedical source, where the target word `observation' has been highlighted:

\enumsentence{The \textbf{observation} of unequal expression leads to a number of questions.}
 
 \noindent In the binary annotation setting of CWI some annotators may rightly consider this term non-complex, whereas others may rightly consider it to be complex. Whilst the meaning of the word is reasonably clear to someone with scientific training, the context in which it is used is unfamiliar for a lay reader and will likely lead to them considering it complex. In our new LCP setting, we are able to ask annotators to mark the word on a scale from very easy to very difficult. Each user can give their subjective interpretation on this scale indicating how difficult they found the word. Whilst annotators will inevitably disagree (some finding it more or less difficult), this is captured and quantified as part of our annotations, with a word of this type likely to lead to a medium complexity value.
 
 LCP is useful as part of the wider task of lexical simplification \cite{Devlin1998}, where it can be used to both identify candidate words for simplification \cite{shardlow-2013-comparison} and rank potential words as replacements \cite{paetzold-specia-2017-lexical}. LCP is also relevant to the field of readability assessment, where knowing the proportion of complex words in a text helps to identify the overall complexity of the text \cite{Dale1948}.
 
 This paper presents SemEval-2021 Task 1: Lexical Complexity Prediction. In this task we developed a new dataset for complexity prediction based on the previously published CompLex dataset. Our dataset covers 10,800 instances spanning 3 genres and containing unigrams and bigrams as targets for complexity prediction. We solicited participants in our task and released a trial, training and test split in accordance with the SemEval schedule. We accepted submissions in two separate Sub-tasks, the first being single words only and the second taking single words and multi-word expressions (modelled by our bigrams). In total 55 teams participated across the two Sub-tasks.
 
 The rest of this paper is structured as folllows: In Section~\ref{RT} we discuss the previous two iterations of the CWI task. In Section~\ref{data}, we present the CompLex 2.0 dataset that we have used for our task, including the methodology we used to produce trial, test and training splits. In Section~\ref{results}, we show the results of the participating systems and compare the features that were used by each system. We finally discuss the nature of LCP in Section~\ref{discussion} and give concluding remarks in Section~\ref{conclusion}
 




\section{Related Tasks}
\label{RT}

\paragraph{CWI 2016 at SemEval} The CWI shared task was organized at SemEval 2016 \cite{paetzold-specia:2016:SemEval1}. The CWI 2016 organizers introduced a new CWI dataset and reported the results of 42 CWI systems developed by 21 teams. Words in their dataset were considered complex if they were difficult to understand for non-native English speakers according to a binary labelling protocol. A word was considered complex if at least one of the annotators found it to be difficult. The training dataset consisted of 2,237 instances, each labelled by 20 annotators and the test dataset had 88,221 instances, each labelled by 1 annotator \cite{paetzold-specia:2016:SemEval1}. 

The participating systems leveraged lexical features \cite{choubey-pateria:2016:SemEval,bingel-schluter-martinezalonso:2016:SemEval,quijada-medero:2016:SemEval} and word embeddings \cite{kuru:2016:SemEval,sp-kumar-kp:2016:SemEval,nat:2016:SemEval}, as well as finding that frequency features, such as those taken from Wikipedia \cite{konkol:2016:SemEval,wrobel:2016:SemEval} were useful. Systems used binary classifiers such as SVMs \cite{kuru:2016:SemEval,sp-kumar-kp:2016:SemEval,choubey-pateria:2016:SemEval}, Decision Trees \cite{choubey-pateria:2016:SemEval,quijada-medero:2016:SemEval,malmasi-dras-zampieri:2016:SemEval}, Random Forests \cite{ronzano-EtAl:2016:SemEval,brooke-uitdenbogerd-baldwin:2016:SemEval,zampieri-tan-vangenabith:2016:SemEval,mukherjee-EtAl:2016:SemEval} and threshold-based metrics \cite{kauchak:2016:SemEval,wrobel:2016:SemEval} to predict the complexity labels. The winning system made use of threshold-based methods and features extracted from Simple Wikipedia \cite{paetzold-specia:2016:SemEval2}.

A post-competition analysis \cite{zampieri-EtAl:2017:NLPTEA} with oracle and ensemble methods showed that most systems performed poorly due mostly to the way in which the data was annotated and the the small size of the training dataset. 

\paragraph{CWI 2018 at BEA} The second CWI Shared Task was organized at the BEA workshop 2018 \cite{stajner-EtAl:2018:BEA}. Unlike the first task, this second task had two objectives. The first objective was the binary complex or non-complex classification of target words. The second objective was regression or probabilistic classification in which 13 teams were asked to assign the probability of a target word being considered complex by a set of language learners. A major difference in this second task was that datasets of differing genres: (TEXT GENRES) as well as English, German and Spanish datasets for monolingual speakers and a French dataset for multilingual speakers were provided \cite{stajner-EtAl:2018:BEA}.

Similar to 2016, systems made use of a variety of lexical features including word length \cite{syspaper10,syspaper6,syspaper9,syspaper4,syspaper2,syspaper5}, frequency \cite{syspaper6,syspaper11,syspaper2,syspaper5}, N-gram features \cite{syspaper7,syspaper3,syspaper4,syspaper2,syspaper1} and word embeddings \cite{syspaper6,syspaper9,syspaper11,syspaper1}. A variety of classifiers were used ranging from traditional machine learning classifiers \cite{syspaper7,syspaper3,syspaper9}, to Neural Networks \cite{syspaper6,syspaper11}. The winning system made use of Adaboost with WordNet features, POS tags, dependency parsing relations and psycholinguistic features \cite{syspaper7}.

\section{Data}
\label{data}

We previously reported on the annotation of the CompLex dataset \cite{shardlow-etal-2020-complex} (hereafter referred to as CompLex 1.0), in which we annotated around 10,000 instances for lexical complexity using the Figure Eight platform. The instances spanned three genres: \textbf{Europarl}, taken from the proceedings of the European Parliament \cite{koehn2005europarl}; \textbf{The Bible}, taken from an electronic distribution of the World English Bible translation \cite{christodouloupoulos2015massively} and \textbf{Biomedical} literature, taken from the CRAFT corpus \cite{bada2012concept}. We limited our annotations to focus only on nouns and multi-word expressions following a Noun-Noun or Adjective-Noun pattern, using the POS tagger from Stanford CoreNLP \cite{manning-etal-2014-stanford} to identify these patterns.

Whilst these annotations allowed us to report on the dataset and to show some trends, the overall quality of the annotations we received was poor and we ended up discarding a large number of the annotations. For CompLex 1.0 we retained only instances with four or more annotations and the low number of annotations (average number of annotators = 7) led to the overall dataset being less reliable than initially expected

For the Shared Task we chose to boost the number of annotations on the same data as used for CompLex 1.0 using Amazon's Mechanical Turk platform. We requested a further 10 annotations on each data instance bringing up the average number of annotators per instance. Annotators were presented with the same task layout as in the annotation of CompLex 1.0 and we defined the Likert Scale points as previously:

\begin{description}

\item[Very Easy:] Words which were very familiar to an annotator.
\item[Easy:] Words with which an annotator was aware of the meaning.
\item[Neutral:] A word which was neither difficult nor easy.
\item[Difficult:] Words which an annotator was unclear of the meaning, but may have been able to infer the meaning from the sentence.
\item[Very Difficult:] Words that an annotator had never seen before, or were very unclear.

\end{description}

\noindent These annotations were aggregated with the retained annotations of CompLex 1.0 to give our new dataset, CompLex 2.0, covering 10,800 instances across single and multi-words and across 3 genres.

The features that make our corpus distinct from other corpora which focus on the CWI and LCP tasks are described below:

\begin{description}
 \item[Continuous Annotations:] We have annotated our data using a 5-point Likert Scale. Each instance has been annotated multiple times and we have taken the mean average of these annotations as the label for each data instance. To calculate this average we converted the Likert Scale points to a continuous scale as follows: Very Easy $\rightarrow$ 0, Easy $\rightarrow$ 0.25, Neutral $\rightarrow$ 0.5, Difficult $\rightarrow$ 0.75, Very Difficult $\rightarrow$ 1.0.
 
 \item[Contextual Annotations:] Each instance in the corpus is presented with its enclosing sentence as context. This ensures that the sense of a word can be identified when assigning it a complexity value. Whereas previous work has reannotated the data from the CWI--2018 shared task with word senses \cite{strohmaier-etal-2020-secoda}, we do not make explicit sense distinctions between our tokens, instead leaving this task up to participants.
 
 \item[Repeated Token Instances:] We provide more than one context for each token (up to a maximum of five contexts per genre). These words were annotated separately during annotation, with the expectation that tokens in different contexts would receive differing complexity values. This deliberately penalises systems that do not take the context of a word into account.
 
 \item[Multi-word Expressions:] In our corpus we have provided 1,800 instances of multi-word expressions (split across our 3 sub-corpora). Each MWE is modelled as a Noun-Noun  or Adjective-Noun pattern followed by any POS tag which is not a noun. This avoids selecting the first portion of complex noun phrases. There is no guarantee that these will correspond to true MWEs that take on a meaning beyond the sum of their parts, and further investigation into the types of MWEs present in the corpus would be informative.
 
 \item[Aggregated Annotations:] By aggregating the Likert scale labels we have generated crowd-sourced complexity labels for each instance in our corpus. We are assuming that, although there is inevitably some noise in any large annotation project (and especially so in crowd-sourcing), this will even out in the averaging process to give a mean value reflecting the appropriate complexity for each instance. By taking the mean average we are assuming uni-modal distributions in our annotations.
 
 \item[Varied Genres:] We have selected for diverse genres as mentioned above. Previous CWI datasets have focused on informal text such as Wikipedia and multi-genre text, such as news. By focusing on specific texts we force systems to learn generalised complexity annotations that are appropriate in a cross-genre setting.
 
\end{description}

\noindent We have presented summary statistics for CompLex 2.0 in Table \ref{tab:complex2.0}. In total, 5,617 unique words are split across 10,800 contexts, with an average complexity across our entire dataset of 0.321. Each genre has 3,600 contexts, with each split between 3,000 single words and 600 multi-word expressions. Whereas single words are slightly below the average complexity of the dataset at 0.302, multi-word expressions are much more complex at 0.419, indicating that annotators found these more difficult to understand. Similarly Europarl and the Bible were less complex than the corpus average, whereas the Biomedical articles were more complex. The number of unique tokens varies from one genre to another as the tokens were selected at random and discarded if there were already more than 5 occurrences of the given token already in the dataset. This stochastic selection process led to a varied dataset with some tokens only having one context, whereas others have as many as five in a given genre. On average each token has around 2 contexts.

\begin{table*}[ht!]
    \centering
    \scalebox{.92}{
    \begin{tabular}{llccc}
        \hline
        \bf Subset & \bf Genre & \bf Contexts & \bf Unique Tokens  & \bf Average Complexity \\\hline
        \multirow{4}{*}{All} &        
         \bf Total     &  \bf 10,800 &  \bf 5,617 &  \bf 0.321 \\ 
        & Europarl & 3,600  & 2,227 & 0.303 \\
        & Biomed   & 3,600  & 1,904 & 0.353 \\
        & Bible    & 3,600  & 1,934 & 0.307 \\\hline
        \multirow{4}{*}{Single} &        
         \bf  Total      &  \bf 9,000  &  \bf 4,129 &  \bf 0.302 \\ 
        & Europarl & 3,000  & 1,725 & 0.286 \\
        & Biomed   & 3,000  & 1,388 & 0.325 \\
        & Bible    & 3,000  & 1,462 & 0.293 \\\hline
        \multirow{4}{*}{MWE} &        
         \bf  Total     &  \bf 1,800 &  \bf 1,488  &  \bf 0.419 \\ 
        & Europarl & 600   & 502   & 0.388 \\
        & Biomed   & 600   & 516   & 0.491 \\
        & Bible    & 600   & 472   & 0.377 \\
        \hline
    \end{tabular}
    }
    \caption{The statistics for CompLex  2.0.}
    \label{tab:complex2.0}
\end{table*}

\section{Data Splits}
In order to run the shared task we partitioned our dataset into Trial, Train and Test splits and distributed these according to the SemEval schedule. A criticism of previous CWI shared tasks is that the training data did not accurately reflect the distribution of instances in the testing data. We sought to avoid this by stratifying our selection process for a number of factors. The first factor we considered was genre. We ensured that an even number of instances from each genre was present in each split. We also stratified for complexity, ensuring that each split had a similar distribution of complexities. Finally we also stratified the splits by token, ensuring that multiple instances containing the same token occurred in only one split. This last criterion  ensures that systems do not overfit to the test data by learning the complexities of specific tokens in the training data.

Performing a robust stratification of a dataset according to multiple features is a non-trivial optimisation problem. We solved this by first grouping all instances in a genre by token and sorting these groups by the complexity of the least complex instance in the group. For each genre, we passed through this sorted list and for each set of 20 groups we put the first group in the trial set, the next two groups in the test set and the remaining 17 groups in the training data. This allowed us to get a rough 5-85-10 split between trial, training and test data. The trial and training data were released in this ordered format, however to prevent systems from guessing the labels based on the data ordering we randomised the order of the instances in the test data prior to release. The splits that we used for the Shared Task are available via GitHub\footnote{\url{https://github.com/MMU-TDMLab/CompLex}}. 

Table \ref{tab:splits} presents statistics on each split in our data, where it can be seen that we were able to achieve a roughly even split between genres across the trial, train and test data.

\begin{table}[ht!]
    \centering
    \scalebox{.92}{
    \begin{tabular}{llccc}
        \hline
        \bf Subset & \bf Genre & \bf Trial & \bf Train  & \bf Test \\\hline
        \multirow{4}{*}{All} &        
         \bf Total     &  \bf  520 & \bf 9179  & \bf 1101  \\ 
        & Europarl & 180 & 3010 & 410 \\
        & Biomed   & 168 & 3090 & 342 \\
        & Bible    & 172 & 3079 & 349 \\\hline
        \multirow{4}{*}{Single} &        
         \bf  Total      &  421 \bf  & 7662 \bf  & 917 \bf  \\ 
        & Europarl & 143 & 2512 & 345 \\
        & Biomed   & 135 & 2576 & 289 \\
        & Bible    & 143 & 2574 & 283 \\\hline
        \multirow{4}{*}{MWE} &        
         \bf  Total     & 99 \bf  & 1517 \bf  & 184 \bf  \\ 
        & Europarl & 37 & 498 & 65 \\
        & Biomed   & 33 & 514 & 53 \\
        & Bible    & 29 & 505 & 66 \\
        \hline
    \end{tabular}
    }
    \caption{The Trial, Train and Test splits that were used as part of the shared task.}
    \label{tab:splits}
\end{table}

\section{Results}\label{results}

The full results of our task can be seen in Appendix~\ref{apx:results}. We had 55 teams participate in our 2 Sub-tasks, with 19 participating in Sub-task 1 only, 1 participating in Sub-task 2 only and 36 participating in both Sub-tasks. We  have used Pearson's correlation for our final ranking of participants, but we have also included other metrics that are appropriate for evaluating continuous and ranked data and provided secondary rankings of these.

Sub-task 1 asked participants to assign complexity values to each of the single words instances in our corpus. For Sub-task 2, we asked participants to submit results on both single words and MWEs. We did not rank participants on MWE-only submissions due to the relatively small number of MWEs in our corpus (184 in the test set).

The metrics we chose for ranking were as follows:
\begin{description}
 \item[Pearson's Correlation:] We chose this metric as our primary method of ranking as it is well known and understood, especially in the context of  evaluating systems with continuous outputs. Pearson's correlation is robust to changes in scale and measures how the input variables change with each other.
 \item[Spearman's Rank:] This metric does not consider the values output by a system, or in the test labels, only the order of those labels. It was chosen as a secondary metric as it is more robust to outliers than Pearson's correlation.
 \item[Mean Absolute Error (MAE):] Typically used for the evaluation of regression tasks, we included MAE as it gives an indication of how close the predicted labels were to the gold labels for our task. 
 \item[Mean Squared Error (MSE):] There is little difference in the calculation of MSE vs. MAE, however we also include this metric for  completeness.
 \item[R2:] This measures the proportion of variance of the original labels captured by the predicted labels. It is possible to do well on all the  other metrics, yet do  poorly on R2 if a system produces annotations with a different distribution than those in the original labels. 
\end{description}

In Table \ref{tab:top10} we show the scores of the top 10 systems across our 2 Sub-tasks according to Pearson's Correlation. We have only reported on Pearson's correlation and R2 in these tables, but the full results with all metrics are available in Appendix \ref{apx:results}. We have included a Frequency Baseline produced using log-frequency from the Google Web1T and linear regression, which was beaten by the majority of our systems. From these results we can see that systems were able to attain reasonably high scores on our dataset, with the winning systems reporting Pearson's Correlation of 0.7886 for Sub-task 1 and 0.8612 for Sub-task 2, as well as high R2 scores of 0.6210 for Sub-task 1 and 0.7389 for Sub-task 2. The rankings remained stable across Spearman's rank, MAE and MSE, with some small variations. Scores were generally higher on Sub-task 2 than on Sub-task 1, and this is likely to be because of the different groups of token-types (single words and MWEs). MWEs are known to be more complex than single words and so this fact may have implictly helped systems to better model the variance of complexities between the two groups.

\begin{table}[!t]
\centering
\scalebox{.95}{
\begin{tabular}{lcc}
\hline
\multirow{2}{*}{\bf Team} & \multicolumn{2}{c}{\bf Task 1} \\
 & \bf Pearson  & \bf R2 \\\hline
JUST BLUE & 0.7886 (1)   & 0.6172 (2) \\
DeepBlueAI      & 0.7882 (2)   & 0.6210 (1) \\
Alejandro Mosquera & 0.7790 (3)   & 0.6062 (3) \\
Andi   & 0.7782 (4)   & 0.6036 (4) \\
CS-UM6P      & 0.7779 (5)   & 0.3813 (47)\\
tuqa            & 0.7772 (6)   & 0.5771 (12)\\
OCHADAI-KYOTO         & 0.7772 (7)   & 0.6015 (5) \\
BigGreen           & 0.7749 (8)   & 0.5983 (6) \\
CSECU-DSC        & 0.7716 (9)   & 0.5909 (8) \\
IA PUCP          & 0.7704 (10)  & 0.5929 (7) \\
\it Frequency Baseline & 0.5287 & 0.2779\\
\hline
 & \multicolumn{2}{c}{\bf Task 2} \\\hline
DeepBlueAI       & 0.8612 (1)  & 0.7389 (1)   \\
rg\_pa           & 0.8575 (2)  & 0.7035 (5)   \\
xiang\_wen\_tian & 0.8571 (3)  & 0.7012 (7)   \\
andi\_gpu        & 0.8543 (4)  & 0.7055 (4)   \\
ren\_wo\_xing    & 0.8541 (5)  & 0.6967 (8)   \\
Andi    & 0.8506 (6)  & 0.7107 (2)   \\
CS-UM6P       & 0.8489 (7)  & 0.6380 (17)  \\
OCHADAI-KYOTO           & 0.8438 (8)  & 0.7103 (3)   \\
LAST              & 0.8417 (9)  & 0.7030 (6)   \\
KFU          & 0.8406 (10) & 0.6967 (9)   \\
\it Frequency Baseline & 0.6571 & 0.4030\\
\hline
\end{tabular}
}
\caption{The top 10 systems for each task according to Pearson's correlation. We have also included R2 score to help interpret the former. For full rankings, see Appendix \ref{apx:results} \label{tab:top10}}
\end{table}

\section{Participating Systems}

In this section we have analysed the participating systems in our task. System Description papers were submitted by 32 teams. In the subsections below, we have first given brief summaries of some of the top systems according to Pearson's correlation for each task for which we had a description. We then discuss the features used across different systems, as well as the approaches to the task that different teams chose to take. We have prepared a comprehensive table comparing the features and approaches of all systems for which we have the relevant information in Appendix \ref{apx:systems}.

\subsection{System Summaries}

\paragraph{DeepBlueAI:} This system attained the highest Pearson's Correlation on Sub-task 2 and the second highest Pearson's Correlation on Sub-task 1. It also attained the highest R2 score across both tasks. The system used an ensemble of pre-trained language models fine-tuned for the task with Pseudo Labelling, Data Augmentation, Stacked Training Models and Multi-Sample Dropout. The data was encoded for the transformer models using the genre and token as a query string and the given context as a supplementary input. 

\paragraph{JUST BLUE:} This system attained the highest Pearson's Correlation for Sub-task 1. The system did not participate in Sub-task 2.  This system makes use of an ensemble of BERT  and RoBERTa. Separate models are fine-tuned for context and token prediction and these are weighted 20-80 respectively. The average of the BERT models and RoBERTa models is taken to give a final score.

\paragraph{RG\_PA:} This system attained the second highest Pearson's Correlation for Sub-task 2. The system uses a fine-tuned RoBERTa model and boosts the training data for the second task by identifying similar examples from the single-word portion of the dataset to train the multi-word classifier. They use an ensemble of RoBERTa models in their final classification, averaging the outputs to enhance performance.

\paragraph{Alejandro Mosquera:}
This system attained the third highest Pearson's Correlation for Sub-task 1. The system used a feature-based approach, incorporating length, frequency, semantic features from WordNet and sentence level readability features. These were passed through a Gradient Boosted Regression.

\paragraph{Andi:} This system attained the fourth highest Pearson's Correlation for Sub-task 1. They combine a traditional feature based approach with features from pre-trained language models. They use psycholinguistic features, as well as GLoVE and Word2Vec Embeddings. They also take features from an ensemble of Language models: BERT, RoBERTa, ELECTRA, ALBERT, DeBERTa. All features are passed through Gradient Boosted Regression to give the final output score.

\paragraph{CS-UM6P:} This system attained the fifth highest Pearson's Correlation for Sub-task 1 and the seventh highest Pearson's Correlation for Sub-task 2. The system uses BERT and RoBERTa and encodes the context and token for the language models to learn from. Interestingly, whilst this system scored highly for Pearson's correlation the R2 metric is much lower on both Sub-tasks. This may indicate the presence of significant outliers in the system's output.

\paragraph{OCHADAI-KYOTO:} This system attained the seventh highest Pearson's Correlation on Sub-task 1 and the eight highest Pearson's Correlation on  Sub-task 2. The system used a fine-tuned BERT and RoBERTa model with the token and context encoded. They employed multiple training strategies to boost performance.

\subsection{Approaches}
There are three main types of systems that were submitted to our task. In line with the state of the art in modern NLP, these can be categorised as: Feature-based systems, Deep Learning Systems and Systems which use a hybrid of the former two approaches. Although Deep Learning Based systems have attained the highest Pearson's Correlation on both Sub-tasks, occupying the first two places in each task, Feature based systems are not far behind, attaining the third and fourth spots on Sub-task 1 with a similar score to the top systems. We have described each approach as applied to our task below.

Feature-based systems use a variety of features known to be useful for lexical complexity. In particular, lexical frequency and word length feature heavily with many different ways of calculating these metrics such as looking at various corpora and investigating syllable or morpheme length. Psycholinguistic features which model people's perception of words are understandably popular for this task as complexity is a perceived phenomenon. Semantic features taken from WordNet modelling the sense of the word and it's ambiguity or abstractness have been used widely, as well as sentence level features aiming to model the context around the target words. Some systems chose to identify named entities, as these may be innately more difficult for a reader. Word inclusion lists were also a popular feature, denoting whether a word was found on a given list of easy to read vocabulary. Finally, word embeddings are a popular feature, coming from static resources such as GLoVE or Word2Vec, but also being derived through the use of Transformer models such as BERT, RoBERTa, XLNet or GPT-2,  which provide context dependent embeddings suitable for our task.

These features are passed through a regression system, with Gradient Boosted Regression and Random Forest Regression being two popular approaches amongst participants for this task. Both apply scale invariance meaning that less pre-processing of inputs is necessary. 

Deep Learning Based systems invariably rely on a pre-trained language model and fine-tune this using transfer learning to attain strong scores on the task.  BERT and RoBERTa were used widely in our task, with some participants also opting for ALBERT, ERNIE, or other such language models. To prepare data for these language models, most participants following this approach concatenated the token with the context, separated by a special token $(\langle SEP\rangle)$. The Language Model was then trained and the embedding of the $\langle CLS\rangle$ token extracted and passed through a further fine-tuned network for complexity prediction. Adaptations to this methodology include applying training strategies such as adversarial training, multi-task learning, dummy annotation generation and capsule networks.

Finally, hybrid approaches use a mixture of Deep Learning by fine-tuning a neural network alongside feature-based approaches. The features may be concatenated to the input embeddings, or may be concatenated at the output prior to further training. Whilst this strategy appears to be the best of both worlds, uniting linguistic knowledge with the power of pre-trained language models, the hybrid systems do not tend to perform as well as either feature based or deep learning systems.

\subsection{MWEs}
For Sub-task 2 we asked participants to submit both predictions for single words and multi-words from our corpus. We hoped this would encourage participants to consider models that adapted single word lexical complexity to multi-word lexical complexity. We observed a number of strategies that participants employed to create the annotations for this secondary portion of our data.

For systems that employed a deep learning approach, it was relatively simple to incorporate MWEs as part of their training procedure. These systems encoded the input using a query and context, separated by a $\langle SEP\rangle$ token. The number of tokens prior to the $\langle SEP\rangle$ token did not matter and either one or two tokens could be placed there to handle single and multi-word instances simultaneously.

However, feature based systems could not employ this trick and needed to devise more imaginative strategies for handling MWEs. Some systems handled them by averaging the features of both tokens in the MWE, or by predicting scores for each token and then averaging these scores. Other systems doubled their feature space for MWEs and trained a new model which took the features of both words into account.

\section{Discussion}\label{discussion}


In this paper we have posited the new task of Lexical Complexity Prediction. This builds on previous work on Complex Word Identification, specifically by providing annotations which are continuous rather than binary or probabilistic as in previous tasks.  Additionally, we provided a dataset with annotations in context, covering three diverse genres and incorporating MWEs, as well as single tokens. We have moved towards this task, rather than rerunning another CWI task as the outputs of the models are more useful for a diverse range of follow-on tasks. For example, whereas CWI is particularly useful as a preprocessing step for Lexical simplification (identifying which words should be transformed), LCP may also be useful for readability assessment or as a rich feature in other downstream NLP tasks. A continuous annotation allows a ranking to be given over words, rather than binary categories, meaning that we can not only tell whether a word is likely to be difficult for a reader, but also how difficult that word is likely to be. If a system requires binary complexity (as in the case of lexical simplification) it is easy to transform our continuous complexity values into a binary value by placing a threshold on the complexity scale. The value of the threshold to be selected will likely depend on the target audience, with more competent speakers requiring a higher threshold. When selecting a threshold, the categories we used for annotation should be taken into account, so for example a threshold of 0.5 would indicate all words that were rated as neutral or above.

To create our annotated dataset, we employed crowdsourcing with a Likert scale and aggregated the categorical judgments on this scale to give a continuous annotation. It should be noted that this is not the same as giving a truly continuous judgment (i.e., asking each annotator to give a value between 0 and 1). We selected this protocol as the Likert Scale is familiar to annotators and allows them to select according to defined points (we provided the definitions given earlier at annotation time). The annotation points that we gave were  not intended to give an even distribution of annotations and it was our expectation that most words would be familiar to some degree, falling in the very easy or easy categories. We pre-selected for harder words to ensure that there were also words in the difficult and very difficult categories. As such, the corpus we have presented is not designed to be representative of the distribution of words across the English language. To create such a corpus, one would need to annotate all words according to our scale with no filtering. The general distribution of annotations in our corpus is  towards the easier end of the Likert scale.

A criticism of the approach we have employed is that it allows for subjectivity in the annotation process. Certainly one annotator's perception of complexity will be different to another's. Giving fixed values of complexity for every word will not reflect the specific difficulties that one reader, or one reader group will face. The annotations we have provided are averaged values of the annotations given by our annotators, we chose to keep all instances, rather than filtering out those where annotators gave a wide spread of complexity annotations. Further work may be undertaken to give interesting insights into the nature of subjectivity in annotations. For example, some words may be rated as easy or difficult by all annotators, whereas others may receive both easy and difficult annotations, indicating that the perceived complexity of the instance is more subjective. We did not make the individual annotations available as part of the shared task data, to encourage systems to focus primarily on the prediction of complexity.


An issue with the previous shared tasks is that scores were typically low and that systems tended to struggle to beat reasonable baselines, such as those based on lexical frequency. We were pleased to see that systems participating in our task returned scores that indicated that they had learnt to model the problem well (Pearson's Correlation of 0.7886 on Task  1 and 0.8612 on Task 2). MWEs are typically more complex than single words and it may be the case that these exhibited a lower variance, and were thus easier to predict for the systems. The strong Pearson's Correlation is backed up by a high R2 score (0.6172 for Task 1 and 0.7389 for Task 2), which indicates that the variance in the data is captured accurately by the models' predictions. These models strongly outperformed a reasonable baseline based on  word frequency as shown in Table~\ref{tab:top10}.

Whilst we have chosen in this report to rank systems based on their score on Pearson's correlation, giving a final ranking over all systems, it should be noted that there is very little variation in score between the top systems and all other systems. For Task 1 there are 0.0182 points of Pearson's Correlation separating the systems at ranks 1 and 10. For Task 2 a similar difference of 0.021 points of Pearson's Correlation separates the systems at ranks 1 and 10. These are small differences and it may be the case that had we selected a different random split in our dataset this would have led to a different ordering in our results \cite{gorman-bedrick-2019-need,sogaard2020we}. This is not unique to our task and is something for the SemEval community to ruminate on as the focus of NLP tasks continues to move towards better evaluation rather than better systems.

An analysis of the systems that participated in our task showed that there was little variation between Deep Learning approaches and Feature Based approaches, although Deep Learning approaches ultimately attained the highest scores on our data. Generally the Deep Learning and Feature Based approaches are interleaved in our results table, showing that both approaches are still relevant for LCP. One factor that did appear to affect system output was the inclusion of context, whether that was in a deep learning setting or a feature based setting. Systems which reported using no context appeared to perform worse in the overall rankings. Another feature that may have helped performance is the inclusion of previous CWI datasets \cite{yimam-etal-2017-cwig3g2,maddela2018word}. We were aware of these  when developing the corpus and attempted to make our data sufficiently distinct in style to prevent direct reuse of these resources.

A limitation of our task is that it focuses solely on LCP for the English Language. Previous CWI shared tasks \cite{stajner-EtAl:2018:BEA} and simplification efforts \cite{saggion2015making,aluisio-gasperin-2010-fostering} have focused on languages other than English and we hope to extend this task in the future to other languages.

\section{Conclusion}\label{conclusion}

We have presented the SemEval-2021 Task 1 on Lexical Complexity Prediction. We developed a new dataset focusing on continuous annotations in context across three genres. We solicited participants via SemEval and 55 teams submitted results across our two Sub-tasks. We have shown the results of these systems and discussed the factors that helped systems to perform well. We have analysed all the systems that participated and categorised their findings to help future researchers understand which approaches are suitable for LCP. 



\bibliography{anthology,CWI,Matt}
\bibliographystyle{acl_natbib}
\onecolumn
\appendix
\section{Full Results} \label{apx:results}

\begin{table*}[!h]
\centering
\scalebox{0.85}{
\begin{tabular}{lccccc}
\hline
\bf Team            & \bf Pearson  & \bf Spearman     & \bf MAE         & \bf MSE         & \bf R2           \\ \hline
JUST BLUE               & 0.7886 (1)   & 0.7369 (2)   & 0.0609 (54) & 0.0062 (53) & 0.6172 (2)   \\
DeepBlueAI              & 0.7882 (2)   & 0.7425 (1)   & 0.0610 (53) & 0.0061 (54) & 0.6210 (1)   \\
Alejandro Mosquera      & 0.7790 (3)   & 0.7355 (5)   & 0.0619 (50) & 0.0064 (52) & 0.6062 (3)   \\
Andi                    & 0.7782 (4)   & 0.7287 (12)  & 0.0637 (40) & 0.0064 (51) & 0.6036 (4)   \\
CS-UM6P                 & 0.7779 (5)   & 0.7366 (3)   & 0.0803 (8)  & 0.0100 (8)  & 0.3813 (47)  \\
tuqa                    & 0.7772 (6)   & 0.7344 (6)   & 0.0635 (43) & 0.0068 (43) & 0.5771 (12)  \\
OCHADAI-KYOTO           & 0.7772 (7)   & 0.7313 (9)   & 0.0617 (52) & 0.0065 (50) & 0.6015 (5)   \\
BigGreen                & 0.7749 (8)   & 0.7294 (11)  & 0.0629 (48) & 0.0065 (49) & 0.5983 (6)   \\
CSECU-DSG               & 0.7716 (9)   & 0.7326 (8)   & 0.0632 (46) & 0.0066 (47) & 0.5909 (8)   \\
ia pucp                 & 0.7704 (10)  & 0.7361 (4)   & 0.0618 (51) & 0.0066 (48) & 0.5929 (7)   \\
CLP                     & 0.7692 (11)  & 0.7336 (7)   & 0.0631 (47) & 0.0067 (46) & 0.5854 (9)   \\
ess                     & 0.7656 (12)  & 0.7308 (10)  & 0.0635 (44) & 0.0069 (39) & 0.5747 (16)  \\
ismail2022              & 0.7653 (13)  & 0.7245 (16)  & 0.0641 (37) & 0.0069 (42) & 0.5766 (13)  \\
andi\_gpu               & 0.7651 (14)  & 0.7275 (13)  & 0.0629 (49) & 0.0068 (44) & 0.5810 (11)  \\
TUDA-CCL                & 0.7649 (15)  & 0.7164 (26)  & 0.0643 (34) & 0.0067 (45) & 0.5846 (10)  \\
rg\_pa                  & 0.7628 (16)  & 0.7251 (15)  & 0.0634 (45) & 0.0069 (40) & 0.5749 (15)  \\
ren\_wo\_xing           & 0.7618 (17)  & 0.7229 (18)  & 0.0639 (39) & 0.0069 (37) & 0.5715 (18)  \\
CLULEX                  & 0.7588 (18)  & 0.7089 (30)  & 0.0649 (29) & 0.0069 (41) & 0.5753 (14)  \\
acccb                   & 0.7586 (19)  & 0.7207 (20)  & 0.0635 (42) & 0.0069 (38) & 0.5730 (17)  \\
jiu\_mo\_zhi            & 0.7584 (20)  & 0.7175 (24)  & 0.0635 (41) & 0.0070 (33) & 0.5691 (22)  \\
Eslam93                 & 0.7577 (21)  & 0.7224 (19)  & 0.0640 (38) & 0.0070 (31) & 0.5648 (24)  \\
archer                  & 0.7561 (22)  & 0.7067 (33)  & 0.0641 (36) & 0.0069 (36) & 0.5707 (19)  \\
Cambridge               & 0.7556 (23)  & 0.7105 (29)  & 0.0646 (32) & 0.0070 (35) & 0.5705 (20)  \\
eee                     & 0.7553 (24)  & 0.7203 (21)  & 0.0673 (20) & 0.0078 (20) & 0.5181 (35)  \\
CompNA                  & 0.7552 (25)  & 0.7153 (27)  & 0.0641 (35) & 0.0070 (34) & 0.5701 (21)  \\
LAST                    & 0.7534 (26)  & 0.6988 (37)  & 0.0652 (27) & 0.0070 (32) & 0.5652 (23)  \\
Stanford MLab           & 0.7533 (27)  & 0.7044 (34)  & 0.0653 (25) & 0.0071 (29) & 0.5615 (26)  \\
mau\_lih                & 0.7513 (28)  & 0.7263 (14)  & 0.0645 (33) & 0.0071 (27) & 0.5587 (28)  \\
IITK@LCP                & 0.7511 (29)  & 0.7167 (25)  & 0.0654 (24) & 0.0071 (28) & 0.5598 (27)  \\
cognience               & 0.7510 (30)  & 0.7193 (22)  & 0.0652 (28) & 0.0071 (30) & 0.5625 (25)  \\
qnamqj                  & 0.7509 (31)  & 0.7086 (31)  & 0.0649 (30) & 0.0072 (26) & 0.5536 (29)  \\
feras1515               & 0.7503 (32)  & 0.7180 (23)  & 0.0652 (26) & 0.0073 (23) & 0.5477 (32)  \\
eslam                   & 0.7482 (33)  & 0.7237 (17)  & 0.0649 (31) & 0.0072 (25) & 0.5525 (30)  \\
RS\_GV                   & 0.7478 (34)  & 0.7077 (32)  & 0.0698 (17) & 0.0079 (18) & 0.5144 (37)  \\
LucasHub                & 0.7434 (35)  & 0.6995 (36)  & 0.0658 (22) & 0.0073 (24) & 0.5486 (31)  \\
LRL\_NC                  & 0.7402 (36)  & 0.7013 (35)  & 0.0661 (21) & 0.0074 (22) & 0.5440 (33)  \\
Manchester Metropolitan & 0.7389 (37)  & 0.7135 (28)  & 0.0656 (23) & 0.0074 (21) & 0.5398 (34)  \\
UPB                     & 0.7340 (38)  & 0.6785 (40)  & 0.0699 (16) & 0.0079 (16) & 0.5098 (39)  \\
KFU                     & 0.7201 (39)  & 0.6899 (39)  & 0.0687 (18) & 0.0079 (17) & 0.5109 (38)  \\
PolyU CBS-Comp          & 0.7188 (40)  & 0.6935 (38)  & 0.0682 (19) & 0.0078 (19) & 0.5162 (36)  \\
LCP\_RIT                & 0.7086 (41)  & 0.6535 (43)  & 0.0716 (15) & 0.0086 (15) & 0.4695 (40)  \\
UNBNLP                  & 0.6953 (42)  & 0.6544 (42)  & 0.0716 (14) & 0.0089 (13) & 0.4495 (42)  \\
chenshi                 & 0.6951 (43)  & 0.6532 (44)  & 0.0740 (11) & 0.0091 (11) & 0.4366 (44)  \\
UTFPR                   & 0.6875 (44)  & 0.6588 (41)  & 0.0735 (13) & 0.0088 (14) & 0.4577 (41)  \\
Katildakat            & 0.6715 (45)  & 0.6454 (46)  & 0.0756 (10) & 0.0096 (10) & 0.4060 (45)  \\
jct                     & 0.6663 (46)  & 0.6457 (45)  & 0.0736 (12) & 0.0091 (12) & 0.4402 (43)  \\
LECCE                   & 0.6452 (47)  & 0.6405 (47)  & 0.0772 (9)  & 0.0096 (9)  & 0.4046 (46)  \\
S3003183                & 0.5834 (48)  & 0.5437 (48)  & 0.0804 (7)  & 0.0110 (7)  & 0.3182 (48)  \\
\it Frequency Baseline 	&0.5287 &	0.5263 	&0.0870 &	0.0136 &	0.2779  \\
C3SL                    & 0.4598 (49)  & 0.3983 (50)  & 0.0866 (6)  & 0.0130 (6)  & 0.1989 (49)  \\
SINAI                   & 0.4428 (50)  & 0.3961 (51)  & 0.0875 (5)  & 0.0131 (5)  & 0.1930 (50)  \\
ProjectLIN513           & 0.3884 (51)  & 0.4316 (49)  & 0.1019 (4)  & 0.0159 (4)  & 0.0198 (51)  \\
glitterosu              & 0.1807 (52)  & 0.1516 (52)  & 0.1024 (3)  & 0.0194 (3)  & -0.2016 (52) \\
PyGuajo                 & 0.0971 (53)  & 0.1440 (53)  & 0.1166 (2)  & 0.0338 (2)  & -1.0861 (53) \\
RACAI                   & -0.0272 (54) & -0.0268 (54) & 0.2777 (1)  & 0.1270 (1)  & -6.8449 (54) \\
\hline
\end{tabular}
}
\caption{Sub-task 1: Results and rank (in brackets) in terms of Pearson, Spearman, MAE, MSE, and R2.}
\end{table*}

\begin{table*}[!h]
\centering
\scalebox{0.85}{
\begin{tabular}{lccccc}
\hline
\bf Team            & \bf Pearson      & \bf Spearman     & \bf MAE         & \bf MSE         & \bf R2           \\ \hline
DeepBlueAI              & 0.8612 (1)  & 0.8526 (3)  & 0.0616 (37) & 0.0063 (37) & 0.7389 (1)   \\
rg\_pa                  & 0.8575 (2)  & 0.8529 (2)  & 0.0672 (33) & 0.0072 (33) & 0.7035 (5)   \\
xiang\_wen\_tian        & 0.8571 (3)  & 0.8548 (1)  & 0.0675 (32) & 0.0072 (31) & 0.7012 (7)   \\
andi\_gpu               & 0.8543 (4)  & 0.8448 (5)  & 0.0664 (35) & 0.0071 (34) & 0.7055 (4)   \\
ren\_wo\_xing           & 0.8541 (5)  & 0.8473 (4)  & 0.0677 (30) & 0.0073 (30) & 0.6967 (8)   \\
Andi                    & 0.8506 (6)  & 0.8381 (7)  & 0.0667 (34) & 0.0070 (36) & 0.7107 (2)   \\
CS-UM6P                 & 0.8489 (7)  & 0.8406 (6)  & 0.0760 (20) & 0.0087 (21) & 0.6380 (17)  \\
OCHADAI-KYOTO           & 0.8438 (8)  & 0.8285 (10) & 0.0660 (36) & 0.0070 (35) & 0.7103 (3)   \\
LAST                    & 0.8417 (9)  & 0.8299 (9)  & 0.0677 (31) & 0.0072 (32) & 0.7030 (6)   \\
KFU                     & 0.8406 (10) & 0.8337 (8)  & 0.0686 (28) & 0.0073 (29) & 0.6967 (9)   \\
jiu\_mo\_zhi            & 0.8355 (11) & 0.8277 (11) & 0.0710 (25) & 0.0083 (23) & 0.6560 (15)  \\
CSECU-DSG               & 0.8311 (12) & 0.8153 (17) & 0.0678 (29) & 0.0077 (27) & 0.6825 (11)  \\
acccb                   & 0.8310 (13) & 0.8157 (15) & 0.0697 (27) & 0.0076 (28) & 0.6850 (10)  \\
Stanford MLab           & 0.8280 (14) & 0.8124 (18) & 0.0711 (24) & 0.0080 (24) & 0.6671 (14)  \\
IITK@LCP                & 0.8277 (15) & 0.8228 (12) & 0.0811 (11) & 0.0098 (15) & 0.5949 (23)  \\
qnamqj                  & 0.8246 (16) & 0.8227 (13) & 0.0787 (17) & 0.0094 (17) & 0.6097 (21)  \\
LRL\_NC                  & 0.8244 (17) & 0.8156 (16) & 0.0702 (26) & 0.0079 (26) & 0.6737 (12)  \\
mau\_lih                & 0.8234 (18) & 0.8211 (14) & 0.0790 (15) & 0.0096 (16) & 0.6042 (22)  \\
TUDA-CCL                & 0.8190 (19) & 0.8091 (19) & 0.0711 (23) & 0.0080 (25) & 0.6677 (13)  \\
Alejandro Mosquera      & 0.8093 (20) & 0.8017 (20) & 0.0731 (22) & 0.0084 (22) & 0.6519 (16)  \\
LucasHub                & 0.8000 (21) & 0.7797 (25) & 0.0754 (21) & 0.0089 (20) & 0.6323 (18)  \\
UPB                     & 0.7962 (22) & 0.7988 (21) & 0.0788 (16) & 0.0099 (14) & 0.5917 (24)  \\
CompNA                  & 0.7931 (23) & 0.7800 (24) & 0.0783 (19) & 0.0093 (18) & 0.6160 (20)  \\
justglowing             & 0.7902 (24) & 0.7851 (23) & 0.0786 (18) & 0.0092 (19) & 0.6169 (19)  \\
BigGreen                & 0.7898 (25) & 0.7769 (26) & 0.0903 (6)  & 0.0124 (6)  & 0.4858 (32)  \\
Katildakat            & 0.7848 (26) & 0.7869 (22) & 0.0807 (12) & 0.0101 (13) & 0.5816 (25)  \\
Manchester Metropolitan & 0.7611 (27) & 0.7711 (27) & 0.0806 (13) & 0.0102 (12) & 0.5770 (26)  \\
UTFPR                   & 0.7601 (28) & 0.7504 (28) & 0.0817 (10) & 0.0102 (11) & 0.5754 (27)  \\
UNBNLP                  & 0.7515 (29) & 0.7420 (30) & 0.0802 (14) & 0.0106 (10) & 0.5623 (28)  \\
chenshi                 & 0.7500 (30) & 0.7497 (29) & 0.0867 (7)  & 0.0112 (8)  & 0.5365 (30)  \\
PolyU CBS-Comp          & 0.7416 (31) & 0.7222 (32) & 0.0839 (9)  & 0.0109 (9)  & 0.5473 (29)  \\
cognience               & 0.7232 (32) & 0.7301 (31) & 0.0851 (8)  & 0.0117 (7)  & 0.5144 (31)  \\
\it Frequency Baseline &	0.6571 	&0.6345 &	0.0924 &	0.0140& 	0.4030\\
C3SL                    & 0.3941 (33) & 0.3675 (34) & 0.1145 (4)  & 0.0206 (4)  & 0.1470 (34)  \\
PyGuajo                 & 0.3931 (34) & 0.3902 (33) & 0.1132 (5)  & 0.0205 (5)  & 0.1488 (33)  \\
SINAI                   & 0.3197 (35) & 0.3508 (35) & 0.1217 (2)  & 0.0243 (2)  & -0.0062 (36) \\
LECCE                   & 0.2821 (36) & 0.3138 (36) & 0.1202 (3)  & 0.0226 (3)  & 0.0624 (35)  \\
glitterosu              & 0.1860 (37) & 0.1316 (37) & 0.1332 (1)  & 0.0255 (1)  & -0.0564 (37) \\
\hline
\end{tabular}
}
\caption{Sub-task 2: Results and rank (in brackets) in terms of Pearson, Spearman, MAE, MSE, and R2.}
\end{table*}
\clearpage

\section{System Features}
\label{apx:systems}

\begin{table*}[!ht]
    \centering
    \scalebox{0.85}{
    \begin{tabular}{l p{8cm} p{5cm}}
    \hline
       \bf Team  &  \multicolumn{1}{c}{\bf Features} & \multicolumn{1}{c}{\bf Classification Approach} \\\hline
Alejandro Mosquera  & Length, Frequency, Semantic, Sentence & Gradient Boosted Regression \\
Andi    & Psycholinguistic, Glove, Word2Vec, ConceptNet NumberBatch, BERT, RoBERTa, ELECTRA, ALBERT, DeBERTa & Ridge Regression, Gradient Boosted Regression\\
Archer    & Length, Frequency, Psycholinguistic, Scrabble Score, Word Inclusion, Semantic & Random Forest Regression, Gradient Boosted Regression \\
BigGreen    & Length, Semantic, Glove, Elmo, InferSent, Phonetic, Frequency, POS & Gradient Boosted Regression, BERT \\
C3SL    & Sent2Vec & Multi-layer Perceptron \\
Cambridge    & Frequency, Syntactic, Length & BERT, Random Forest Regression\\
CLULEX    & Frequency, POS, Named Entities, Word Inclusion, Sentence, Bert & Decision Tree \\
CompNA    & Length, Semantic, Glove, Word Inclusion, & Decision Tree Ensemble\\
CS-UM6P    & Token and Context Encoded & BERT, RoBERTa \\
CSECU-DSG    & Token and Context Encoded & BERT, RoBERTa\\
DeepBlueAI    & Token and Context Encoded & BERT, ALBERT, RoBERTa, ERNIE\\
IA PUCP    & Sentence, POS, N-gram Frequency, RoBERTa, XLNet, BERT & Gradient Boosted Regression \\
IITK@LCP    & Electra + Glove & Linear Regression, Support Vector Machine \\
JCT    & POS, Frequency, BERT, Cluster Features & Gradient Boosted Regression\\
JUST BLUE    & Token Encoded and Context Encoded & Average of Weighted Bert and Roberta \\
Katildakat    &BERT, Length, BERT-score, Frequency, Semantic,  & Linear Regression, Multi-layer Perceptron\\
LAST    & Frequency, Psycholinguiistic, Sentence, Bigram Association & Gradient Boosted Regression \\



LCP-RIT    & Length, Frequency, Character N-Grams, Psycholinguistic, POS & Random forest Regressor\\
LRL\_NC    & Frequency, Semantic, Laanguage Model, Psycholinguistic, Word Inclusion & Random forest regressor \\
Hub    & TF-IDF, Context Encoded & RoBERTa, Inception \\
Manchester Metropolitan    & Frequency, Psycholinguistic, Length, Embeddings & CNN \\
OCHADAI-KYOTO    & Token and Context Encoded  & BERT, RoBERTa\\
PolyU CBS-Comp    & Frequency, Length, Capitalisation, POS, Embeddings, BERT, GPT-2 & Gradient Boosted Regression\\
RG\_PA    & Context Encoded & RoBERTa\\
RS\_GV   & GLoVE, ELMo, BERT, Flair, Readability, Length, Frequency, Semantic, Psycholinguistic, Morphological, Word Inclusion, Named Entity & Feed-Forward Neural Network \\
S3003183    & Token and Context Encoded, Glove, Word2Vec  & BERT, Feed Forward NN \\
SINAI    & Frequency, Length, POS, Semantic & Random Forest Regressor \\
Stanford MLab    & Glove, Length, POS, Named Entity & Gradient Boosted Regression \\
TUDA-CCL    & Linguistic, Semantic, Embeddings, Psycholinguistic, Frequencies, Word Inclusion & Gradient Boosted Regression\\
UNBNLP    & Length, Frequency, Character-Level-Encoder, BERT & Neural Network, Support Vector Machine\\
UPB    & Transformers, Word Embeddings, Character Wmbeddings, Length, Psycholinguistic & BERT, RoBERTa, Regression \\
UTFPR    & Frequency, Length, Semantic, Bert Embedding & Support Vector Machine\\
        \hline
    \end{tabular}
    }
    \caption{Part 2. Systems that participated and submitted a paper, the features and classification approaches they employed.}
    \label{tab:system_descriptions2}
\end{table*}

\end{document}